\documentclass[10pt,letterpaper]{article}

\usepackage{ccn}
\usepackage{pslatex}
\usepackage{apacite}
\usepackage{graphicx}

\title{The role of object-centric representations, guided attention, and external memory on generalizing visual relations}
 
\author{{\large \bf Guillermo Puebla (guillermo.puebla@cenia.cl)} \\
  National Center for Artificial Intelligence, Vicuña Mackenna 4860 \\
  Macul, Chile
  \AND {\large \bf Jeffrey S. Bowers (j.bowers@bristol.ac.uk)} \\
  School of Psychological Science, University of Bristol, 12a Priory Road \\
  Bristol, United Kingdom}

\begin{document}

\maketitle

\section{Abstract}
{
\bf
Visual reasoning is a long-term goal of vision research. In the last decade, several works have attempted to apply deep neural networks (DNNs) to the task of learning visual relations from images, with modest results in terms of the generalization of the relations learned. In recent years, several innovations in DNNs have been developed in order to enable learning abstract relation from images. In this work, we systematically evaluate a series of DNNs that integrate mechanism such as slot attention, recurrently guided attention, and external memory, in the simplest possible visual reasoning task: deciding whether two objects are the same or different. We found that, although some models performed better than others in generalizing the same-different relation to specific types of images, no model was able to generalize this relation across the board. We conclude that abstract visual reasoning remains largely an unresolved challenge for DNNs.
}
\begin{quote}
\small
\textbf{Keywords:} 
visual reasoning; relational generalization; deep neural networks; guided attention; external memory
\end{quote}

\section{Introduction}
Detecting relations is one of the fundamental operations of the visual system. This allows us to form a coherent representation of the environment as sets of relations between objects \cite{vo2019reading}. It is also the basis of robust object recognition, because representing an object as a set of relations between parts frees us from recognizing it solely on the basis of its superficial features \cite{biederman1987recognition}. Furthermore, representing relations between entities forms the basis of the kind of reasoning abilities that set us apart from other species \cite{gentner2021learning}. Given this predominant role across different forms of visual processing, several researchers have attempted to apply deep neural networks to visual reasoning, in particular to the same-different task (i.e., classifying an image with two objects as an example of the categories "same" or "different"). This previous research found that, in contrast with earlier machine learning models, convolutional neural networks (CNNs) can learn to classify images with abstract shapes as same or different \cite<e.g.,>[]{messina2021solving, funke2021five}. However, \citeA{puebla2022can} showed that the representations learned by these models are highly specific: when trained in task \#1 of the synthetic visual reasoning test \cite<SVRT, >[see Original condition Figure 1]{fleuret2011comparing}, CNNs tended to classify correctly images that were superficially similar to the ones they were trained on (e.g., Irregular or Regular conditions in Figure 1) and misclassify images that illustrated the same relation but were more superficially dissimilar (e.g., Lines or Arrows conditions). In the meantime, a number of new DNNs have introduced architectural innovations targeted at achieving relational visual relational reasoning. In this work, we test relational generalization of the same-different task on these models. 

\begin{figure}[t!]
\begin{center}
\includegraphics[width=\linewidth]{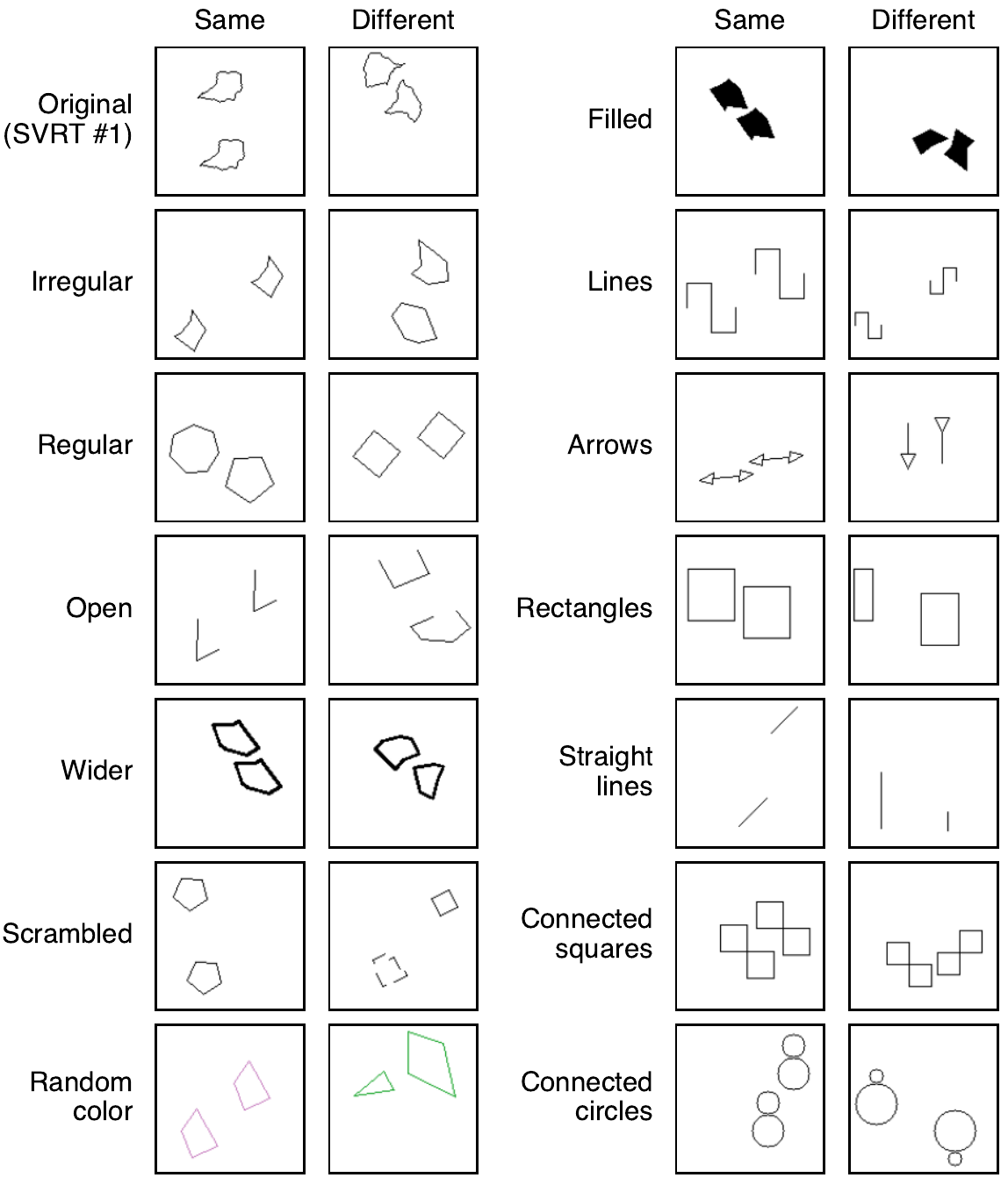}
\end{center}
\caption{Example images of all datasets tested.} 
\label{figure-1}
\end{figure}

\section{Models}

\subsubsection{ResNet50 \cite{he2016deep}}
We included this model as a baseline deep CNN. 

\subsubsection{Slot Attention \cite{locatello2020object}}
This model segregates objects in an image through a key-value attention mechanism that assigns different parts of the image to competing slots. 

\subsubsection{Recurrent vision transformer \cite<RViT,>[]{messina2022recurrent}}
This model applies a standard vision transformer encoder recurrently, that is the model takes as input its own output, for a number of 4 processing steps. 

\subsubsection{Emergent symbol binding network \cite<ESBN,>[]{webb2021emergent}}
This model consist of a recurrent neural network augmented with a key-value external memory. This model aims to bind its memory values (direct representation of the input) and keys (inferred representation of the input's role in the sequence).  

\subsubsection{Guided Attention Model for (visual) Reasoning \cite<GAMR,>[]{vaishnav2023gamr}}
This model is composed of three modules. An encoder builds a representation of the input. A recurrent controller guides an attention mechanism to select relevant object representations and write them into an external memory. A graph neural network module computes relations between the objects stored in the external memory.

\subsubsection{Object-Centric Recurrent Attention \cite<OCRA,>[]{adeli2022brain}}
This model consist of a recurrent encoder that controls an attention window that trades of the area it covers by its resolution. At the same time, a recurrent decoder control a write window that modifies a reconstruction output at each time step. The encoder feeds a two-layer capsule neural network that predicts a class label. 

\subsection{Methods}
We trained 10 runs of all the models in task \#1 of SVRT until reaching a validation accuracy of approximately 99\%. This dataset consist of 28,000 128$\times$128 RGB images. We tested all the models in all 14 datasets illustrated in Figure 1 (5,600 images per dataset), with the exception that OCRA was not tested on the Random colors dataset since this model takes only grey scale images as an input.

\begin{figure}[t!]
\begin{center}
\includegraphics[width=\linewidth]{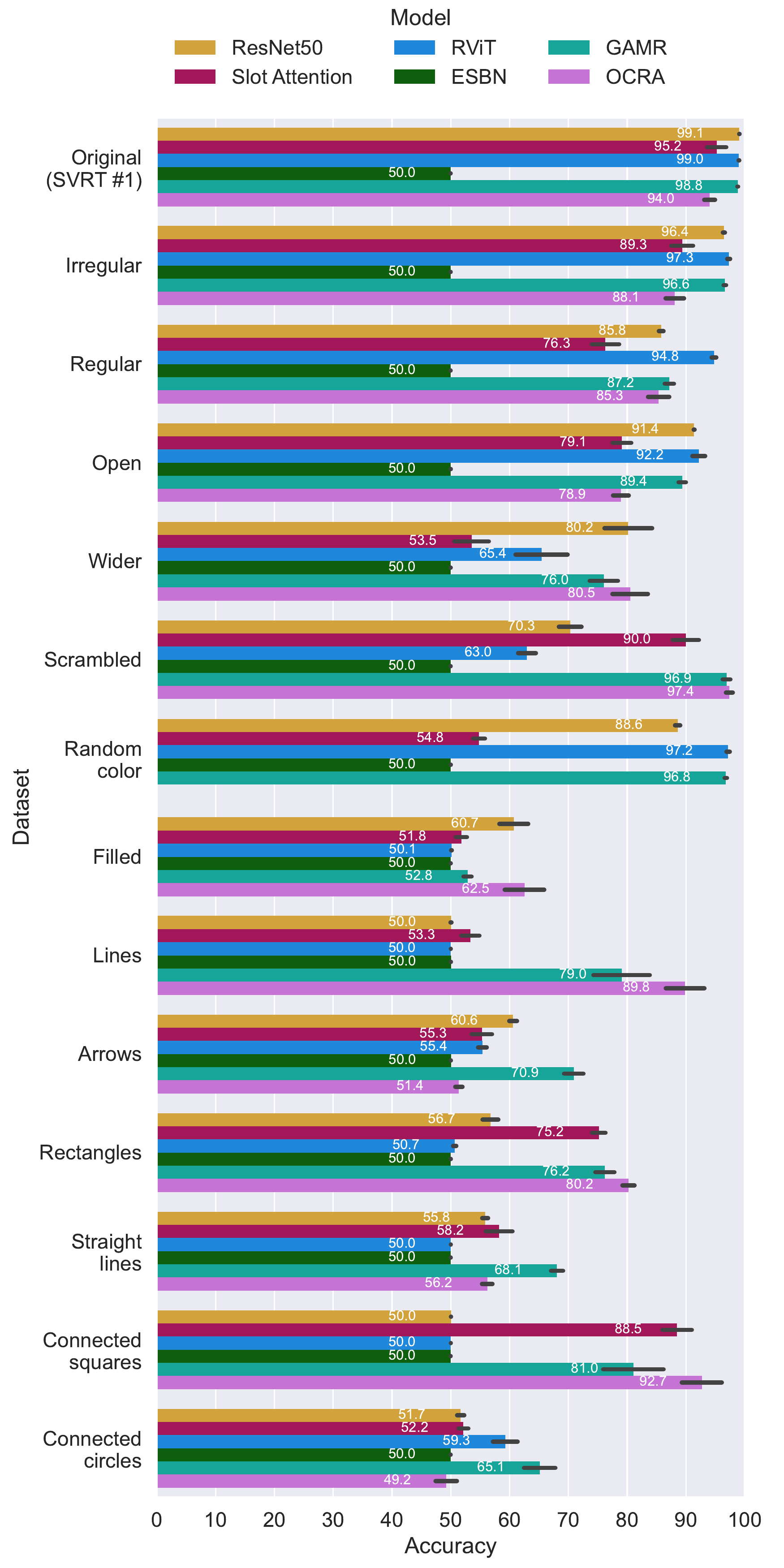}
\end{center}
\caption{Mean accuracy on 10 runs of each model per condition. Error bars are standard errors of the mean.} 
\label{figure-2}
\end{figure}

\subsection{Results and discussion}

As can be seen in Figure 1, all models achieved high accuracy in the test split of task \#1 of SVRT except for the ESBN model, which performed at chance in all datasets. Further analysis showed that this model can learn the same-different task only when the objects (presented individually in a sequence of two images) are centered in the image, which severely questions \citeA{webb2021emergent}'s conclusions regarding the relational reasoning capabilities of the model.

Furthermore, Slot Attention, GAMR and OCRA tended to show better generalization on the datasets that were harder for ResNet50. However, no single model achieved high levels of accuracy across all the test datasets, which is what woukd be expected if a model learned an abstract representation of the relations "same" and "different". 

\section{Acknowledgments}

The first author has received funding for this project from the National Center for Artificial Intelligence CENIA FB210017, Basal ANID.

The second author has received funding for this project from the European Research Council (ERC) under the European Union’s Horizon 2020 research and innovation programme (grant agreement No 741134) for the second author.

\bibliographystyle{apacite}

\setlength{\bibleftmargin}{.125in}
\setlength{\bibindent}{-\bibleftmargin}

\clearpage
\bibliography{ccn_style}

\end{document}